\documentclass{article}
\PassOptionsToPackage{numbers, compress}{natbib}
\usepackage[dblblindworkshop,final]{neurips_2026}
\workshoptitle{AI Foundations for Power Grids: From Models to Deployment at Scale}
\usepackage[utf8]{inputenc}
\usepackage{subcaption}
\usepackage[T1]{fontenc}
\usepackage{hyperref,url,booktabs,amsfonts,amsmath,amssymb,bm,nicefrac,microtype,xcolor,graphicx}
\usepackage{booktabs}
\usepackage{subcaption}
\usepackage{tabularx}
\usepackage{array}
\usepackage{natbib}

\DeclareMathOperator*{\softmax}{softmax}

\title{Sensor-Aware Joint Data Fusion and State Estimation\\under Sparse Multi-rate Distribution Grid Sensing}
\author{Pooja Algikar\textsuperscript{\rm 1,*},
Stefano Fenu\textsuperscript{\rm 1,*},
Kumarsinh Jhala\textsuperscript{\rm 1},
Siby Plathottam\textsuperscript{\rm 1}
\\
\textsuperscript{\rm 1}Argonne National Laboratory,
Lemont, Illinois, USA
\\
}

\begin{document}
\maketitle

\begin{abstract}
Distribution-system state estimation is a growing necessity for the modern power grid as the addition of local storage and generation add system complexity that an operator or aggregator is not able to directly monitor. Classical methods work well when all system elements are time-synchronized and at least partly observable, but multi-rate monitoring of sparse heterogeneous sensors typically relies on machine learning approaches. 
It is a common practice in the literature to structure this task as a two-stage pipeline, first fusing data from disparate sensors and then feeding the fused pseudo-measurements to a separately trained state estimator. 
We propose a heterogeneous latent graph ODE in which measurement functions and sampling processes are typed and localized on specific buses, paired with a graph-transformer state estimator solving DSSE specifically for multi-rate heterogeneous sensor placements, and demonstrate its effectiveness across multiple IEEE standard topologies as observability degrades.
\end{abstract}

\section*{Introduction}

Most distribution grids were originally designed as passive radial delivery systems, with limited sensing. The underlying assumption of unidirectional flow from substation to load and of slowly varying aggregate demand were baked into the original engineering of these systems, and instrumentation reflected these assumptions. Modern developments have made distribution systems significantly more complex, with distributed grid resources (DGRs) and behind-the-meter storage violating both assumptions. Grid operators now find themselves needing situational awareness on infrastructure that is not instrumented to provide it, with sensor deployments being heterogeneous and fragmentary. Deploying more sensors can alleviate some of these issues, but adoption and installation of novel sensors is slow, expensive, and piecemeal, leaving most distribution grids with a patchwork of sensing modalities varying in sampling frequency, noise characteristics, and measurement modalities.
Operational tasks that hinge on system visibility, such as fault localization, congestion management, hosting capacity assessment, volt/VAR optimization, DGR setpoint dispatch, therefore require state estimation in order to infer bus voltage phasors across the feeder from whatever measurements the deployed sensor mix provides. The result is a growing need for distribution system state estimation (DSSE) methods that are capable of working with multi-rate heterogeneous sensor mixes and that are robust to partial observation. 

Classical methods for state estimation tend to rely on weighted least-squares (WLS) estimators, which assume a synchronized measurement snapshot and a higher degree of observability than is typical in distribution systems. Pseudo-measurements injection can restore observability but bounds accuracy. Snapshot approaches to WLS do not capture system temporal dynamics, and cannot deal with unsynchronized or multi-rate measurements that are common in modern distribution systems. 
A growing literature converges on GNN-based data-driven approaches to state estimation, which exploit the feeder topology directly through message passing along electrical adjacency: GNNs over factor graphs \cite{kundacina2022state}, hypergraph deep statistical solvers with physics-regularized training \citep{habib2023deep}, topology-aware architectures for PMU-unobservable systems \citep{moshtagh2025topology}, and attention-augmented variants \citep{yaniv2027robust}. This work, however, generally uses uniform sampling rates and homogeneous sensor distributions, which increasingly fails to reflect the condition of modern grids as it treats measurement input as a time-aligned snapshot or regularly sampled sequence, and asynchrony must be resolved before the model ever sees the data. The neural ODE literature resolves non-uniform sampling as latent ODE models condition a continuous latent trajectory on irregularly timed observations rather than forcing a discretized grid \citep{chen2018neural} but has not been used for heterogeneous spatial measurements as its application in distribution grids to date treats the feeder as an unstructured multivariate time series \citep{dahale2023latent}. Graph ODE models exist, coupling latent ODE dynamics with GNN vector fields over interaction graphs \citep{huang2020learning}, but these have not been used for heterogeneous sensor modalities and assume homogeneous node types sharing a single emission model and observation space. We extend the latent graph ODE framework with typed encoders, dynamics, and decoders so that the observation structure of a real feeder is represented natively rather than approximated away. 

\section*{Method}

We represent the feeder as a graph $G=(V,E)$ with buses $V=\{1,\dots,N\}$. The state $x_i(t)\in\mathbb{R}^d$ (per-phase voltage, current, real and reactive injection) obeys the AC power-flow equations $s_i(t)=\sum_{j\in\mathcal{N}(i)} v_i(t)(Y_{ij}v_j(t))^{*}$. Each sensor has a location $\ell(s)\in V$ and a class $\tau\in\mathcal{T}=\{\text{PMU},\text{SCADA},\text{smart meter},\text{DER}\}$ determining which subset $\mathcal{C}_\tau$ of a global feature catalog it observes and at what reporting interval. Given $G$, the sensor set, and observations $O$ over $[0,T]$, the task is to estimate $x(t)$ at every bus, including uninstrumented ones. At any instant, the observed set is typically insufficient for WLS.

Het-ODE builds on LG-ODE \citep{huang2020learning}, which infers a per-node initial latent with a graph recognition model, evolves it with a neural ODE, and decodes the trajectory. LG-ODE shares projections, dynamics, and decoder across all nodes, an assumption that fails when nodes are sensors of different classes. Instead our approach retains a common latent space but conditions selected transformations on $\tau(i)$. Each observation is masked to its class catalog and mapped into the shared space by a type-specific adapter, then concatenated with a learned type embedding:
\begin{equation}
  h^{t,0}_i=\sigma\!\big(A^{\tau(i)}(o^t_{i,\mathcal{C}_{\tau(i)}}\odot m^t_{i,\mathcal{C}_{\tau(i)}})\big),\qquad
  \tilde h^{t,0}_i=[\,h^{t,0}_i\,\|\,e_{\tau(i)}\,],
  \label{eq:adapter}
\end{equation}
which aligns heterogeneous feature sets before fusion. For a full derivation see Appendix~\ref{app:method}. The latent state evolves as $z^T_i=z^0_i+\int_0^T g_i(Z^t;\tau)dt$ with shared relation and object functions and a type-specific readout $\dot z^t_i=W^{\tau(i)}_{o}(\cdot)$, so all modalities obey one interaction law while the latent derivative at a PMU bus may be expressed differently from that at a DER bus.

The fused reconstruction $\hat x^{\mathrm{DF}}(t)$ is refined by a graph-transformer estimator emitting a per-bus residual, $\hat x_i(t)=\hat x^{\mathrm{DF}}_i(t)+\Delta_i(t)$ with $\Delta(t)=E_\vartheta([\hat x^{\mathrm{DF}}(t)\|m(t)];G)$ and $m(t)$ the observation mask (Full derivation in Appendix~\ref{app:se}).

\paragraph{Objectives.}
Training proceeds in two stages with different objectives, and we state both explicitly because they differ. The fusion stage minimizes the negative ELBO of the generative process:
\begin{equation}
  \mathcal{L}_{\mathrm{rec}}=-\,\mathbb{E}_{z^0\sim q_\phi}\Big[\textstyle\sum_{i,t}\log p^{\tau(i)}_\theta\big(x_i(t)\mid z^t_i\big)\Big]+\beta_k\,\mathrm{KL}\big[q_\phi(z^0\mid O)\,\big\|\,\mathcal{N}(0,I)\big].
  \label{eq:lrec}
\end{equation}
The estimation stage then optimizes the fusion and estimation parameters jointly under a masked squared error on the \emph{final} estimate,
\begin{equation}
  \mathcal{L}_{\mathrm{est}}=\frac{\sum_{i,t}m_i(t)\,\big(\hat x_i(t)-x_i(t)\big)^2}{\sum_{i,t}m_i(t)}.
  \label{eq:lest}
\end{equation}

\section*{Results}

\paragraph{Dataset.}
We evaluate on three IEEE test feeders: 13-bus (13 buses), 34-bus (36), and 123-bus (130). Exogenous loads are driven by a model fit to real utility AMI profiles and AC power flow is solved in OpenDSS to obtain ground truth states which are then filtered by a candidate sensor assignment. Sensor classes are assigned to buses following a stochastic rule-based algorithm (Appendix~\ref{app:assign}). Each sensor class has a distinct sampling interval (60\,s for PMU and DER, 300\,s for SCADA, 900\,s for smart meters). A field PMU streams at tens of hertz, here we use ``PMU'' to denote the measurement channel set (per-phase voltage and current magnitude and angle) delivered at a 60\,s reporting cadence as a simplification to prevent PMU measurements from dominating the full set of observations. All results use a 60-day dataset generated this way per topology, retaining the first 42 days for training and testing on the last 18. 

\paragraph{Accuracy:}
Table~\ref{tab:pipeline} gives fusion alone against the full pipeline. The estimation stage removes 20\% to 47\% of the fusion residual, and the margin grows with feeder size. Throughout this work, we compare to a naive baseline estimator returning the per-bus historical mean in order to characterize how much of the time-varying spatial and temporal information is getting captured by the model from the available observations.

\begin{table}[t] \centering \small \begin{subtable}[t]{0.58\textwidth} \centering \caption{Per-unit MSE, 60-day chronological holdout. Results are means (standard deviations) across 3 seeds.} \label{tab:pipeline} \begin{tabular}{lccc} \toprule Feeder & DF only & DF + SE \\ \midrule IEEE 13-bus & 0.0283 $(.00135)$ & 0.0226 $(.00165)$  \\ 
IEEE 34-bus & 0.0855 $(.00461)$ & 0.0449 $(.00184)$  \\
IEEE 123-bus & 0.0543 $ (.00331)$ & 0.0329 $(.00217)$   \\ \bottomrule 
\end{tabular} 
\end{subtable} 
\hfill 
\begin{subtable}[t]{0.38\textwidth} \centering \caption{Data fusion model performance across all selected state estimators on IEEE 123.} \label{tab:dfxse} \begin{tabular}{lcc} \toprule State Estimator & Het-ODE & LG-ODE \\ \midrule GAT & 0.1642 & 0.1785 \\ GraphTransformer & \textbf{0.1201} & 0.1608 \\ GCN & 0.1373 & 0.2271 \\ GIN & 0.5123 & 1.1606 \\ \bottomrule \end{tabular} \end{subtable} \end{table}
Ablation of model choice for each pipeline component is given in \ref{tab:dfxse}, showing an empirical increase in performance across topologies for Het-ODE + graph transformer under matched training conditions.

\begin{figure}[t]
  \centering
  \includegraphics[width=0.49\linewidth]{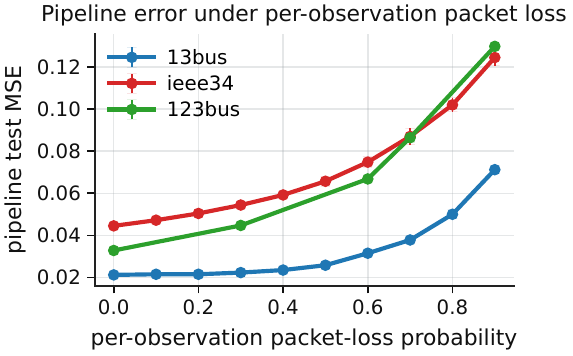}\hfill
  \includegraphics[width=0.49\linewidth]{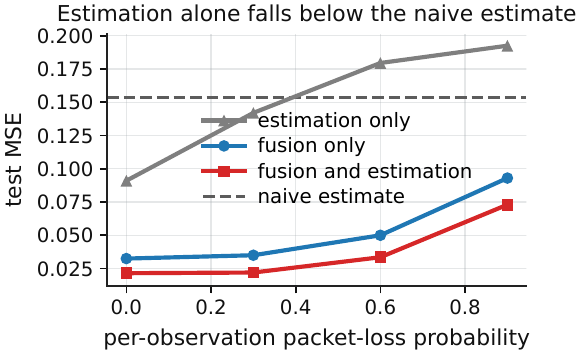}
  \caption{\emph{Left:} Pipeline error under per-observation packet loss.
  Mean $\pm$ standard deviation over three perturbation draws on one trained
  model. \emph{Right:} State-estimation and data-fusion performance on the
  13-bus system under increasing observation missingness.}
  \label{fig:pktloss}
\end{figure}

\paragraph{Degradation:}
Figure~\ref{fig:pktloss} reports state estimation error as each observation is dropped independently with probability $p$. Tested topologies degrade differently, with the difference tracking sensor mix rather than size. The practical reading is that this pipeline tolerates roughly half of all packets going missing before accuracy meaningfully suffers, and that the tolerance is set by how many observations the fast-reporting classes contribute more than by feeder size.


Separating the stages under the same perturbation (Figure~\ref{fig:pktloss}) illustrates the concrete advantage provided by the fusion stage. An estimator operating directly on sparse masked snapshots begins to perform worse than a naive estimator returning the per-bus mean near 60\% loss ($0.180$ against $0.153$), becoming worse than that trivial predictor beyond that level of observability. The full pipeline remains roughly $2\times$ better than a naive estimate or a state estimator alone even at 90\% loss.



\section*{Discussion}

\begin{figure}[t]
    \centering
    \includegraphics[
        width=0.90\linewidth,
        trim={0bp 0bp 0bp 100bp},
        clip
    ]{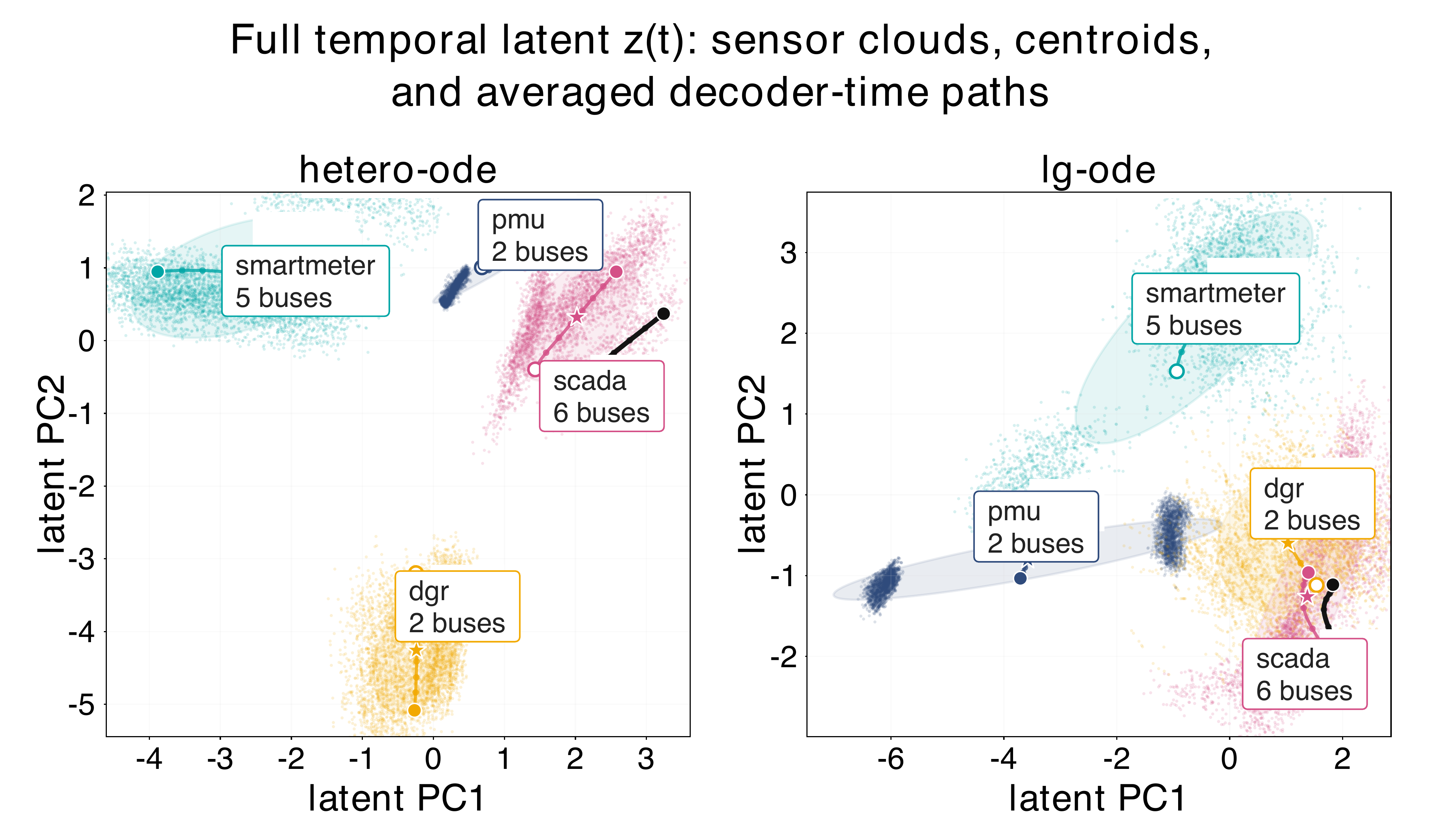}
    \caption{Comparison of the latent representations learned by vanilla
    LG-ODE and Het-ODE. Small dots denote sampled latent states; open and
    filled circles mark the start $\bm{z}^{0}$ and end $\bm{z}^{T}$ of each
    mean trajectory, respectively; stars indicate sensor-type centroids;
    colored curves trace sensor-averaged temporal trajectories; and the
    black curve highlights bus 611. Het-ODE more clearly separates
    physically distinct regimes while maintaining cross-sensor alignment
    for observations associated with similar grid states.}
    \label{fig:latent_comparison}
\end{figure}

Figure~\ref{fig:latent_comparison} compares the posterior initial states $\bm{z}^0$ and the ODE rollouts $\bm{z}^{1:T}$ learned by vanilla LG-ODE and Het-ODE. A desirable heterogeneous representation should separate distinct physical operating conditions while avoiding fragmentation caused solely by sensor identity. Relative to the homogeneous baseline, Het-ODE produces latent representations with clearer organization by physical regime and more coherent continuous-time trajectories. This indicates that the type-conditioned maps absorb modality-specific differences before and during fusion, allowing the shared latent coordinates to represent the underlying grid state more consistently. Results given in \ref{tab:dfxse} indicate that this translates into a performance improvement on heterogeneous sensor data, although we make no claims of a causal relationship and leave the evaluation to future work.

\paragraph{Limitations}

Models are trained per topology on a set of fixed sensor placements, and a topology-independent estimator is future work, as is the co-design of sensor placement with the estimator. 

\section*{Conclusion}

We present a joint approach to DSSE for heterogeneous multi-rate sensing conditions that scales robustly to real feeder system sizes. Our fusion model, Het-ODE, extends latent graph ODEs with type conditioned encoders and attention. We characterize the conditions under which a composed data fusion and state estimation model fails across 3 grid topologies subject to real load changes under different sensing regimes and observability levels. As observability declines, state estimation pipelines develop error that can fall below a trivial baseline and can be rescued by composition with a data fusion model.

\bibliographystyle{plainnat}
\bibliography{references}

\appendix

\section{Attention, posterior, objectives, and hyperparameters}
\label{app:method}

\paragraph{Heterogeneous mutual attention.}
Each observation is a node in a temporal graph whose edges connect temporally and electrically related observations. For an edge from source $s$ to target $t$ with gap $\Delta t_{st}$, a time-aware source representation is formed and attended with type-conditioned projections:
\begin{align}
  \hat h^{l-1}_s &= \sigma\!\big(W^{\tau(s)}_{\mathrm{tr}}[h^{l-1}_s\,\|\,e_{\tau(s)}\,\|\,\Delta t_{st}]\big)+\mathrm{TE}(\Delta t_{st}),\nonumber\\
  q_{t,a} &= W^{\tau(t)}_{q,a}[h^{l-1}_t\,\|\,e_{\tau(t)}],\quad
  k_{s,a} = W^{\tau(s),r_{st}}_{k,a}\hat h^{l-1}_s,\quad
  v_{s,a} = W^{\tau(s),r_{st}}_{v,a}\hat h^{l-1}_s,\nonumber\\
  \alpha_{st,a} &= \softmax_{s\in\mathcal{N}_t}\!\Big(\tfrac{q_{t,a}^{\top}k_{s,a}}{\sqrt{d_a}}\Big),\qquad
  h^{l}_t = h^{l-1}_t+\sigma\!\Big(\big\Vert_{a=1}^{H}\sum_{s\in\mathcal{N}_t}\alpha_{st,a}v_{s,a}\Big),
\end{align}
with $\mathrm{TE}$ a sinusoidal temporal encoding and $r_{st}$ distinguishing self-node temporal edges from cross-node edges. Relative to LG-ODE the new element is conditioning $q,k,v$ on source and target sensor class.

\paragraph{Temporal aggregation and posterior.}
Contextualized observations of each bus are aggregated over its observation times. With mean aggregation $u_i=\frac{1}{T_i}\sum_t h^t_i$; the attention variant replaces this with LG-ODE's sequence attention applied to the sensor-aware representations. The approximate posterior is $q_\phi(z^0_i\mid o_{1:N})=\mathcal{N}(\mu^0_i,\operatorname{diag}((\sigma^0_i)^2))$ with $[\mu^0_i\,\|\,\log\sigma^0_i]=f_\phi(u_i)$. Although $f_\phi$ is shared, its input is sensor-aware because it is produced by the typed adapters and heterogeneous attention.

\paragraph{ODE generator.}
With $e_{ij}\in\{0,1\}^{R}$ the relation-type indicator for edge $(j,i)$:
\begin{equation}
  s^t_i=\sigma(W_z z^t_i),\quad
  m^t_i=\sum_{j\in\mathcal{N}_i}\sum_{r=1}^{R}e_{ij,r}f^r_R([s^t_j\,\|\,s^t_i]),\quad
  \dot z^t_i=W^{\tau(i)}_{o}\big(s^t_i+f_O([s^t_i\,\|\,m^t_i])\big).
\end{equation}

\paragraph{Hyperparameters.}
Latent dimension 16; recognition and ODE hidden dimension 64; one recognition layer and one generator layer; single attention head; dropout 0.2; augmentation dimension 164; two edge types with additive aggregation; observation noise std $10^{-3}$. Optimizer Adam at learning rate $10^{-3}$, global batch size 8, three trajectory samples per step. Fusion checkpoints used for the frozen-model studies are trained to 1200 iterations on 13-bus and IEEE-34 and 600 on 123-bus. The power-flow penalty referenced in Section~5 is gated by a coefficient that defaults to zero and was left at zero for every run reported here.

\section{Estimator architectures}
\label{app:se}
The estimator consumes the per-bus fusion output concatenated with the observation mask, $h^{(0)}_i=W^{(0)}[\hat x^{\mathrm{DF}}_i(t)\|m_i(t)]$, and applies dot-product graph attention over $\mathcal{N}(i)$:
\begin{align}
  \alpha^{(\ell,a)}_{ij}&=\softmax_{j\in\mathcal{N}(i)}\Big(\tfrac{(W^{(\ell,a)}_{Q}h^{(\ell-1)}_i)^{\top}(W^{(\ell,a)}_{K}h^{(\ell-1)}_j)}{\sqrt{d_h}}\Big),\\
  h^{(\ell)}_i&=\sigma^{(\ell)}\Big(W^{(\ell)}_{R}h^{(\ell-1)}_i+\big\Vert_{a=1}^{A}\sum_{j}\alpha^{(\ell,a)}_{ij}W^{(\ell,a)}_{V}h^{(\ell-1)}_j\Big),
\end{align}
with $\sigma^{(\ell)}=\mathrm{ReLU}$ for $\ell<L$; at the output layer head sums are averaged, the activation is the identity, and the emitted quantity is the residual $\Delta_i(t)$. We additionally evaluated GCN, GIN, GAT, and DeepGAT estimator heads on the same interface.

An estimator trained on masked raw snapshots and then fed a dense fused reconstruction is evaluated off its training distribution, and composing one degrades the pipeline below fusion alone. All pipeline numbers in Table~\ref{tab:pipeline} use heads trained through the fusion output. The following are evaluations of the candidate model performance:

\begin{table}[t]\centering\small
\caption{IEEE 13-bus feeder results by SE model and modality}
\label{tab:t1_13bus}
\begin{tabular}{lcccc}
\toprule
SE backbone & se\_only & composed \\
\midrule
GCN & 0.0503  & 0.0025 \\
GIN & 0.0712  & 0.0027  \\
GAT & 0.0655  &  0.0025  \\
DeepGAT & 0.0691  &  0.0025  \\
GraphTF & 0.0071  &  0.0024   \\
\bottomrule
\end{tabular}
\end{table}

\begin{table}[t]\centering\small
\label{tab:34bus}
\caption{IEEE 34 feeder results by SE model and modality}
\begin{tabular}{lccc}
\toprule
SE backbone & se\_only & composed \\
\midrule
GCN & 0.0804 & 0.0002 \\
GIN & 0.1019 & 0.0004  \\
GAT & 0.0714 & 0.0002  \\
DeepGAT & 0.0776 & 0.0001 \\
GraphTF & 0.0020 & 0.0000 \\
\bottomrule
\end{tabular}
\end{table}

\begin{table}[t]\centering\small
\label{tab:123bus}
\caption{IEEE 123 bus results by SE model and modality}
\begin{tabular}{lcccc}
\toprule
SE backbone & se\_only & composed \\
\midrule
GCN & 0.0864 & 0.0003 \\
GIN & 0.0996 & 0.0004 \\
GAT & 0.0838 & 0.0002 \\
DeepGAT & 0.1026 & 0.0003 \\
GraphTF & 0.0043 & 0.0001 \\
\bottomrule
\end{tabular}
\end{table}

\section{Sensor Assignment Algorithm}
\label{app:assign}
The proposed rule-based assignment algorithm places devices according to their functional role and the electrical characteristics of the distribution feeder. First, the OpenDSS model is parsed and solved at a representative loading condition to construct a bus-level feature table containing phase availability, voltage level, local and downstream load, customer count, distance from the source, voltage sensitivity, line loading, hosting-capacity headroom, and equipment indicators. Infeasible connection points—such as source buses, regulator-internal buses, capacitor-only buses, open switches, and buses without valid phase information—are removed before DER placement. Rooftop PV is preferentially assigned to customer and load buses, community PV to three-phase load pockets with adequate hosting headroom, and utility-scale PV to three-phase primary-feeder buses with sufficient voltage and thermal capacity. Candidate buses are ranked using DER-type-specific weighted scores and sampled without replacement, with probabilities proportional to their scores. SCADA placement is primarily deterministic and includes the substation, feeder head, regulators, capacitor banks, reclosers, sectionalizers, switches, tie switches, and major DER interconnection points; optional SCADA locations at important branch nodes are selected using downstream load, voltage sensitivity, line loading, and branch importance. AMI is restricted to positive-load customer buses and excluded from infrastructure-only locations. Under partial AMI coverage, buses are sampled according to customer density and local demand, whereas full coverage assigns an AMI-equivalent measurement to every eligible load bus. Finally, one PMU is fixed at the feeder head or substation, and the remaining PMUs are selected using a score that combines feeder depth, downstream load, voltage sensitivity, branch importance, DER proximity, line loading, and three-phase availability. A minimum feeder-depth separation is enforced to prevent PMU clustering and improve spatial observability. This produces a layered measurement configuration in which SCADA monitors utility infrastructure, AMI represents the customer edge, PMUs provide sparse high-resolution feeder observability, and DERs occupy physically feasible and electrically relevant connection points.

\section{Acknowledgements}
The submitted manuscript has been created by UChicago Argonne, LLC, Operator of Argonne National Laboratory (“Argonne”). Argonne, a U.S. Department of Energy Office of Science laboratory, is operated under Contract No. DE-AC02-06CH11357. This work was supported in part by the U.S. Department of Energy, Office of Electricity. The U.S. Government retains for itself, and others acting on its behalf, a paid-up nonexclusive, irrevocable worldwide license in said article to reproduce, prepare derivative works, distribute copies to the public, and perform publicly and display publicly, by or on behalf of the Government.

\end{document}